\documentclass[runningheads]{llncs}

\usepackage{eccv}

\usepackage{eccvabbrv}

\usepackage{graphicx}
\usepackage{booktabs}
\usepackage{siunitx}
\usepackage{amsmath}
\usepackage{outlines}
\usepackage{comment}
\usepackage{afterpage}
\usepackage{import}
\usepackage{wrapfig}
\usepackage{multirow}

\usepackage[accsupp]{axessibility}  

\usepackage{hyperref}

\usepackage{orcidlink}

\everymath=\expandafter{\the\everymath\displaystyle}
\makeatletter\@ifpackageloaded{underscore}{}{\usepackage[strings]{underscore}}\makeatother

\begin{document}

\title{Minimalist Vision with Freeform Pixels}


\author{
Jeremy Klotz
\and Shree K. Nayar
}

\authorrunning{J.~Klotz \and S.K.~Nayar}

\institute{
Computer Science Department, Columbia University, New York NY, USA
\email{\{jklotz,nayar\}@cs.columbia.edu}
}

\maketitle

\vspace{-0.15in}
\begin{abstract}
A minimalist vision system uses the smallest number of pixels needed to solve a vision task. While traditional cameras use a large grid of square pixels, a minimalist camera uses freeform pixels that can take on arbitrary shapes to increase their information content. 
We show that the hardware of a minimalist camera can be modeled as the first layer of a neural network, where the subsequent layers are used for inference.  
Training the network for any given task yields the shapes of the camera's freeform pixels, each of which is implemented using a photodetector and an optical mask. We have designed minimalist cameras for monitoring indoor spaces (with $8$ pixels), measuring room lighting (with $8$ pixels), and estimating traffic flow (with $8$ pixels). The performance demonstrated by these systems is on par with a traditional camera with orders of magnitude more pixels. Minimalist vision has two major advantages. First, it naturally tends to preserve the privacy of individuals in the scene since the captured information is inadequate for extracting visual details. Second, since the number of  measurements made by a minimalist camera is very small, we show that it can be fully self-powered, i.e., function without an external power supply or a battery. 

\end{abstract}

\afterpage{

\begin{figure}[t]
    \centering
    \def\svgwidth{\linewidth}
\begingroup%
  \makeatletter%
  \providecommand\color[2][]{%
    \errmessage{(Inkscape) Color is used for the text in Inkscape, but the package 'color.sty' is not loaded}%
    \renewcommand\color[2][]{}%
  }%
  \providecommand\transparent[1]{%
    \errmessage{(Inkscape) Transparency is used (non-zero) for the text in Inkscape, but the package 'transparent.sty' is not loaded}%
    \renewcommand\transparent[1]{}%
  }%
  \providecommand\rotatebox[2]{#2}%
  \newcommand*\fsize{\dimexpr\f@size pt\relax}%
  \newcommand*\lineheight[1]{\fontsize{\fsize}{#1\fsize}\selectfont}%
  \ifx\svgwidth\undefined%
    \setlength{\unitlength}{332.53991318bp}%
    \ifx\svgscale\undefined%
      \relax%
    \else%
      \setlength{\unitlength}{\unitlength * \real{\svgscale}}%
    \fi%
  \else%
    \setlength{\unitlength}{\svgwidth}%
  \fi%
  \global\let\svgwidth\undefined%
  \global\let\svgscale\undefined%
  \makeatother%
  \begin{picture}(1,0.35594843)%
    \lineheight{1}%
    \setlength\tabcolsep{0pt}%
    \put(0,0){\includegraphics[width=\unitlength,page=1]{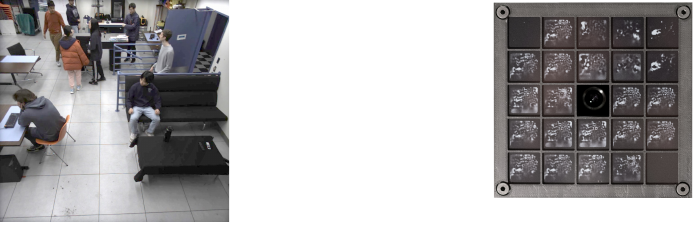}}%
    \put(0.85459838,0.00676611){\color[rgb]{0,0,0}\makebox(0,0)[t]{\lineheight{1.25}\smash{\begin{tabular}[t]{c}(c) Minimalist Camera\end{tabular}}}}%
    \put(0.51862936,0.00676611){\color[rgb]{0,0,0}\makebox(0,0)[t]{\lineheight{1.25}\smash{\begin{tabular}[t]{c}(b) Camera in a Network\end{tabular}}}}%
    \put(0,0){\includegraphics[width=\unitlength,page=2]{teaser.pdf}}%
    \put(0.16425308,0.00682483){\color[rgb]{0,0,0}\makebox(0,0)[t]{\lineheight{1.25}\smash{\begin{tabular}[t]{c}(a) Workspace Monitoring\end{tabular}}}}%
    \put(0,0){\includegraphics[width=\unitlength,page=3]{teaser.pdf}}%
  \end{picture}%
\endgroup%

    \vspace{-0.2in}
    \caption{
    \textbf{Monitoring a workspace with minimalist vision.} 
    (a) The task is to count the number of people, track the occupancy of specified zones, and detect when the door is open. A minimalist vision system, composed of a camera and inference network, can perform such lightweight tasks using just a handful of freeform pixels.  (b) The entire system can be modeled as a single network. Once this network is trained, the first layer specifies the design of a camera, a prototype of which is shown in (c). This system can count the people in the room (with 2 pixels), track the occupancy of each zone (with 2 pixels, each), and detect when the door is open (with 2 pixels). Given the small number of measurements it makes, a minimalist camera can be completely self-powered.
    }
    \label{fig:teaser}
    \vspace{-0.25in}
\end{figure}

}

\section{Why Minimalist  Vision?}

Today, computer vision plays an indispensable role in our everyday lives. It serves as the backbone in a wide gamut of applications ranging from video surveillance and monitoring to autonomous driving and robotics. Broadly speaking, vision applications can be divided into two categories. In one category, the system seeks to infer detailed information about objects and activities in a scene. Examples include object detection and recognition, optical flow estimation and tracking, and 3D reconstruction. The second category of applications involves high-level inferences about the statistics of objects in a scene and the states of an environment. Examples in this realm include monitoring the occupancy of workspaces, the flow of traffic on highways, and the lighting in an urban environment. 

In our work, we are interested in the second category, which we refer to as ``lightweight vision.'' 
We claim that lightweight tasks can be solved not with traditional images, but rather a very small number of measurements, as long as the measurements are rich in information. 

We introduce minimalist vision as an approach to solve lightweight tasks. In the arts, minimalism is a technique that is used to pare down a piece of work to its essential elements. The goal is to ensure that each element used has a purpose. In our context, traditional cameras that are used in virtually all vision systems today capture far more information than needed to solve a lightweight task. Our work seeks to answer two key questions: (a) \textit{Given a task, what is the smallest number of visual measurements needed to achieve a desired performance?}  (b) \textit{How do we construct a camera that produces these measurements?} If we are successful in designing such a  minimalist camera, it would have the following two major benefits:
 
\vspace{0.1in}

\noindent\textbf{Towards Privacy Preservation:} 
When a traditional camera captures an image, it typically reveals far more information about the scene than necessary for the task. For instance, a single image could reveal a person's identity, location, or even intentions. This is a well-known problem that has made the widespread deployment of cameras highly controversial~\cite{ohchrPrivacy}. Since minimalist vision captures the smallest number of measurements for a given task, it is difficult to extract visual details about the scene such as the biometrics of an individual. 
Although we cannot guarantee that privacy will be preserved in all applications, we claim that an inherent feature of our approach is that it tends to preserve privacy.

\vspace{0.1in}
\noindent\textbf{Towards Self-Sustainability:} 
The imaging pipeline of a typical camera involves pixel readout, analog-to-digital (A/D) conversion, signal processing, and transmission. The power consumed by each of these steps, and hence the complete pipeline, is approximately linear in the number of pixels.
Since a minimalist system uses an extremely small number of pixels, it consumes orders of magnitude less power than a typical camera. As a result, a minimalist camera can be designed to function using power harvested from just the light falling upon it, without using an external power supply or a battery. In other words, minimalist cameras can be completely self-sustaining and hence more widely deployed. 
\vspace{0.1in}

To achieve minimalist vision, our key insight is to allow each pixel to have an arbitrary shape. We refer to such a pixel as  a ``freeform pixel.'' We show that a freeform pixel performs a linear projection of the scene, allowing us to model a collection of such pixels as a single layer in a neural network. Thus, a minimalist vision system, comprising both the camera and inference network, can be modeled as one network. For a given task, such as monitoring the indoor workspace in \cref{fig:teaser}(a), we use a video captured from an auxiliary camera to train the network in \cref{fig:teaser}(b). The trained network reveals both the shapes of the freeform pixels and the weights of the inference network. Then, a camera (\cref{fig:teaser}(c)) is fabricated, where each freeform pixel is implemented using an optical mask and a photodetector. In \cref{fig:teaser}(a), the results (people count, door status, and zone occupancy) produced by a camera with only 8 freeform pixels are overlaid on the scene image.

We have conducted extensive synthetic and real experiments that show freeform pixels can solve lightweight tasks using orders of magnitude fewer measurements than a traditional camera. We have used our prototype minimalist camera to demonstrate a variety of tasks: monitoring an indoor space (with $8$ pixels), measuring room lighting (with $8$ pixels), and estimating traffic flow (with $8$ pixels). Finally, we show that our prototype can be powered using just the light falling on it. Under indoor lighting, it can read out and wirelessly transmit the measurements made by $24$ freeform pixels at $30$ frames per second without the use of an external power supply or a battery.

\vspace{-0.1in}
\section{Related Work}
\vspace{-0.06in}

Our work is inspired by Pooj \etal~\cite{poojMinimalistCamera2018}, who introduced the concept of a minimalist camera, where each pixel is a combination of an optical mask and a photodetector. In their work, each mask was handcrafted to solve simple vision tasks such as intrusion detection and object speed estimation. Our work introduces the idea of a freeform pixel that can be automatically designed using training data for any given task. Our key observation is that a camera with freeform pixels can be modeled as the first layer of a neural network. Once the network has been trained, the first layer is used to fabricate the camera, and the rest of the network is used for inference.
Furthermore, we show that minimalist cameras can be fully self-powered, making them more easily deployable than traditional cameras.

Our work is closely related to deep optics, an emerging field that jointly designs optics and software using deep learning~\cite{wetzsteinInferenceArtificialIntelligence2020}.
Sitzmann \etal~\cite{sitzmannEndtoEndOptimizationOptics2018} used this approach to design an optical element for improved image quality.
Subsequently, multiple works have used the approach to design optics for image enhancement~\cite{dunLearnedRotationallySymmetric2020, metzlerDeepOpticsSingleShot2020, sunLearningRank1Diffractive2020} and depth estimation~\cite{haimDepthEstimationSingle2018, changDeepOpticsMonocular2019, wuPhaseCam3DLearningPhase2019}.
A similar approach has been taken to design imaging lenses using differentiable ray tracing~\cite{sunEndtoEndComplexLens2021, liEndtoendLearnedSingle2021, coteDifferentiableLensCompound2023} and differentiable proxy functions~\cite{tsengDifferentiableCompoundOptics2021}.
Tseng \etal~\cite{tsengNeuralNanoopticsHighquality2021} used this technique to design a metasurface lens with improved image quality.
In each of these works, a differentiable model for the camera's optics is incorporated into a neural network, and the optics is designed by training the network for the specific goal. While we follow a similar approach, our motivation is different.
Rather than design optics to enhance image quality or improve task performance, we design cameras that seek to preserve privacy and be self-sustaining by taking the smallest number of measurements.

Prior work has also demonstrated the use of optics  in existing network architectures to reduce the computations required during inference.
Lin \etal~\cite{linAllopticalMachineLearning2018} fabricated an entire image classification network using layers of diffractive optics.
Others have explored hybrid approaches that implement just the first layer of a convolutional network in optics. In \cite{chenASPVisionOptically2016}, angle-sensitive pixels were used to convolve the image with a set of commonly used filters, while in \cite{changHybridOpticalelectronicConvolutional2018}, optical phase masks were used to implement learned filters. The goal of our work is different; it is to minimize the number of visual measurements needed for a task, not to reduce the computations in a trained network.

Duarte \etal~\cite{duarteSinglePixelImagingCompressive2008} proposed a single pixel camera that captures compressive measurements of a scene. While both the single pixel camera and our minimalist camera capture linear projections of the scene, the single pixel camera uses thousands of measurements, acquired in series using a single detector, to reconstruct an image of the scene.
Image and scene reconstruction has also been demonstrated using an image sensor that views the scene through an amplitude mask~\cite{asifFlatCamThinLensless2017}, a phase mask~\cite{boominathanPhlatCamDesignedPhaseMask2020}, and a diffuser~\cite{antipaDiffuserCamLenslessSingleexposure2018}. While all of the above approaches aim to reconstruct an image or 3D scene, the minimalist camera circumvents image reconstruction and seeks to directly solve the task using the smallest number of measurements.

Several works have explored imaging architectures that preserve an individual's privacy while still capturing enough information to perform a task.
Some of them attempt to eliminate visual details related to biometrics in captured images by using low-resolution image sensors~\cite{daiPrivacypreservingActivityRecognition2015} and time-of-flight sensors~\cite{jiaUsingTimeofFlightMeasurements2014}.\footnote{Unrelated to privacy preservation, Torralba \etal~\cite{torralba80MillionTiny2008} demonstrated image classification using a large dataset of very low resolution ($32 \times 32$) images. In our experiments, we compare the performance of our minimalist cameras with low-resolution traditional cameras of different resolutions.}
Others have approached the problem by introducing optical blur~\cite{pittalugaPrivacyPreservingOptics2015, pittalugaPreCapturePrivacySmall2017}, by performing image processing in the analog domain before readout~\cite{tanCANOPICPreDigitalPrivacyEnhancing2020}, or by designing optical elements that preserve the visual feature of interest while eliminating privacy-related details \cite{hinojosaLearningPrivacyPreservingOptics2021, tasneemLearningPhaseMask2022}. Since our approach is to minimize the number of visual measurements, we claim that we implicitly preserve privacy. We demonstrate this via simulations by showing that our freeform pixels are unable to perform face recognition with meaningful accuracy.

It is known that cameras are power-hungry---the image sensor alone can consume hundreds of milliwatts~\cite{likamwaEnergyCharacterizationOptimization2013}. Nayar \etal~\cite{nayarSelfPoweredCameras2015} demonstrated a self-powered camera with $30\times40$ pixels that harvests energy from the light falling on its sensor to read out full images. The harvested energy, however, was insufficient for wireless transmission of the images. Since our minimalist approach results in a small number of pixels, our camera is able to both read out and wirelessly transmit the measurements using just the light falling on it. This is an important feature of our approach; we aim to develop a completely self-sustaining camera that does not need to be tethered and hence can be more widely deployed.

\vspace{-0.1in}
\section{Freeform Pixels}
\label{sec:freeform-pixels}
\vspace{-0.09in}

Since the advent of digital imaging, cameras have used square pixels on a regular grid to record images. While some other pixel tessellations have been suggested in the past~\cite{ben-ezraPenrosePixelsSuperResolution2007, groscheImageSuperResolutionUsing2023}, square pixels have persisted as the standard sensing element. We posit that there exists a large class of vision tasks for which the square pixel forces the camera to capture significantly more measurements than needed.

\begin{figure}[t]
    \centering
    \def\svgwidth{0.95\linewidth}
\begingroup%
  \makeatletter%
  \providecommand\color[2][]{%
    \errmessage{(Inkscape) Color is used for the text in Inkscape, but the package 'color.sty' is not loaded}%
    \renewcommand\color[2][]{}%
  }%
  \providecommand\transparent[1]{%
    \errmessage{(Inkscape) Transparency is used (non-zero) for the text in Inkscape, but the package 'transparent.sty' is not loaded}%
    \renewcommand\transparent[1]{}%
  }%
  \providecommand\rotatebox[2]{#2}%
  \newcommand*\fsize{\dimexpr\f@size pt\relax}%
  \newcommand*\lineheight[1]{\fontsize{\fsize}{#1\fsize}\selectfont}%
  \ifx\svgwidth\undefined%
    \setlength{\unitlength}{302.28861237bp}%
    \ifx\svgscale\undefined%
      \relax%
    \else%
      \setlength{\unitlength}{\unitlength * \real{\svgscale}}%
    \fi%
  \else%
    \setlength{\unitlength}{\svgwidth}%
  \fi%
  \global\let\svgwidth\undefined%
  \global\let\svgscale\undefined%
  \makeatother%
  \begin{picture}(1,0.33426847)%
    \lineheight{1}%
    \setlength\tabcolsep{0pt}%
    \put(0.85996859,0.00744317){\color[rgb]{0,0,0}\makebox(0,0)[t]{\lineheight{0}\smash{\begin{tabular}[t]{c}(c) Mask\end{tabular}}}}%
    \put(0,0){\includegraphics[width=\unitlength,page=1]{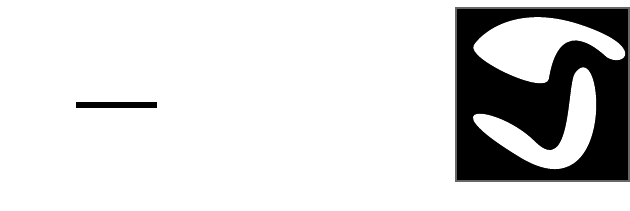}}%
    \put(0.18315661,0.00744317){\color[rgb]{0,0,0}\makebox(0,0)[t]{\lineheight{0}\smash{\begin{tabular}[t]{c}(a) Traditional Pixel\end{tabular}}}}%
    \put(0,0){\includegraphics[width=\unitlength,page=2]{baseline-v-freeform.pdf}}%
    \put(0.53152582,0.00744317){\color[rgb]{0,0,0}\makebox(0,0)[t]{\lineheight{0}\smash{\begin{tabular}[t]{c}(b) Freeform Pixel\end{tabular}}}}%
    \put(0,0){\includegraphics[width=\unitlength,page=3]{baseline-v-freeform.pdf}}%
  \end{picture}%
\endgroup%

    \vspace{-0.05in}
    \caption{\textbf{A freeform pixel can have an arbitrary shape.} (a)~A single pixel in a traditional camera is square and captures light from a small patch in the scene. (b)~A freeform pixel uses a detector and an optical mask to implement any pixel shape. (c)~Example of an optical mask. While this mask is binary, a mask can have any continuous transmittance function. }
    \label{fig:baseline-v-freeform}
    \vspace{-0.2in}
\end{figure}

Consider the traditional camera model shown in \cref{fig:baseline-v-freeform}(a), where a single square pixel (detector) receives light from a scene patch. When the pixel is small, there is a good chance that it will measure information that is not relevant to the task. If the pixel is large, its measurement may include information germane to the task, but it may also be corrupted by unrelated information. In short, if we are interested in capturing the smallest number of measurements for a task, square pixels are almost guaranteed to be the wrong choice.

We propose freeform pixels that can take on an arbitrary shape. As shown in \cref{fig:baseline-v-freeform}(b), a freeform pixel can be implemented by placing an optical mask in front of a photodetector. While we have shown a binary mask in \cref{fig:baseline-v-freeform}(c), each point on the mask can have an arbitrary transmittance. Let us denote the transmittance function as $M(x,y)$, where $0 \le M(x,y) \le 1$. If we assume that the detector is infinitesimally small, then the measurement $p$ produced by it is:
\begin{equation}
    p = \iint_{x,y} I(x,y) \, M(x,y) \, dx \, dy \label{eq:linear-projection} \,.
\end{equation}
Here, $I(x,y)$ is a projection of the 3D scene onto the plane of the mask, where the center of projection is the detector. The above expression shows that, in effect, each freeform pixel performs a linear projection of the scene.

\vspace{-0.1in}
\subsection{Minimalist Camera in a Network}
\vspace{-0.05in}

Since a freeform pixel performs a linear projection, a set of such pixels can be modeled as a single fully-connected layer in a network, without any bias terms.
Based on this observation, we can construct a single network for a task, such as the one in \cref{fig:teaser}(b), that includes freeform pixels and an inference network.
The data for training this network is collected using the training camera shown in \cref{fig:teaser}(c). Once the network is trained, the learned weights of the first layer are used to fabricate (print out) the masks of the freeform pixels. The smallest set of freeform pixels needed to solve a task constitutes a minimalist camera.

It is important to note that the square pixels found in a traditional camera are a special case of freeform pixels.
Thus, if we do not limit the number of pixels used, a minimalist camera can solve any task that a traditional camera can. In general, minimalist vision significantly reduces the number of pixels needed to solve lightweight tasks. As a task becomes more complex, a larger number of freeform pixels would be needed, and the benefits of minimalist vision diminish. For fine-grained tasks (e.g. optical flow or face identification), the number of freeform pixels needed can be expected to approach that of a traditional camera. In short, the use of freeform pixels only serves to dramatically reduce the number of measurements needed to solve a task.

\vspace{-0.1in}
\subsection{Sensor Model} \label{sec:sensor-model}
While a freeform pixel gives us flexibility, it is subjected to a set of physical constraints.
First, the mask transmittance must be positive and cannot be greater than 1.
Second, the detector will have a directional response and a non-zero active area. Finally, the detector will have a limited dynamic range, and its output will include noise. It is important to take all of these factors into consideration when modeling the first layer (freeform pixels) of the network. We now describe the complete sensor model we have developed and how it can be incorporated into the network.

\vspace{-0.1in}
\subsubsection{Optics:}
As shown in \cref{fig:sensor-model}, the detector placed behind the mask is expected to have a response that varies with the direction $\theta$ of the incoming light. We can represent the directional response as a function $d(x,y)$; it acts like a vignetting function with attenuation that increases with $\theta$. Therefore, the light field received by the detector can be modeled as $I(x,y)\,M(x,y)\,d(x,y)$.

In practice, any detector would have a non-zero active area. The effect of this active area can be modeled by blurring  $I(x,y)$ with a kernel $b(x,y)$, the width of which equals that of the active area. The total light energy received by the detector can therefore be expressed as:
\begin{equation}
    p_d = \iint_{x,y} \bigl( I(x,y) * b(x,y) \bigr) \, M(x,y)  \, d(x,y) \, dx \, dy \,. \label{eq:pd}
\end{equation}

As mentioned before, the value of the mask transmittance function $M(x,y)$ must lie between 0 and 1. When we model a pixel as a part of a network, however, it is desirable to let all the trainable parameters be unbounded. To this end, we define $M_t(x,y)$, such that $M(x,y) = \sigma (M_t(x,y))$, where $\sigma$ is the sigmoid function. The trainable parameters are now represented by $M_t(x,y)$, and the corresponding $M(x,y)$ is guaranteed to lie between 0 and 1.

\subsubsection{Detector:}
\begin{figure}[t]
    \centering
    \def\svgwidth{\linewidth}
\begingroup%
  \makeatletter%
  \providecommand\color[2][]{%
    \errmessage{(Inkscape) Color is used for the text in Inkscape, but the package 'color.sty' is not loaded}%
    \renewcommand\color[2][]{}%
  }%
  \providecommand\transparent[1]{%
    \errmessage{(Inkscape) Transparency is used (non-zero) for the text in Inkscape, but the package 'transparent.sty' is not loaded}%
    \renewcommand\transparent[1]{}%
  }%
  \providecommand\rotatebox[2]{#2}%
  \newcommand*\fsize{\dimexpr\f@size pt\relax}%
  \newcommand*\lineheight[1]{\fontsize{\fsize}{#1\fsize}\selectfont}%
  \ifx\svgwidth\undefined%
    \setlength{\unitlength}{305.18076324bp}%
    \ifx\svgscale\undefined%
      \relax%
    \else%
      \setlength{\unitlength}{\unitlength * \real{\svgscale}}%
    \fi%
  \else%
    \setlength{\unitlength}{\svgwidth}%
  \fi%
  \global\let\svgwidth\undefined%
  \global\let\svgscale\undefined%
  \makeatother%
  \begin{picture}(1,0.34153604)%
    \lineheight{1}%
    \setlength\tabcolsep{0pt}%
    \put(0,0){\includegraphics[width=\unitlength,page=1]{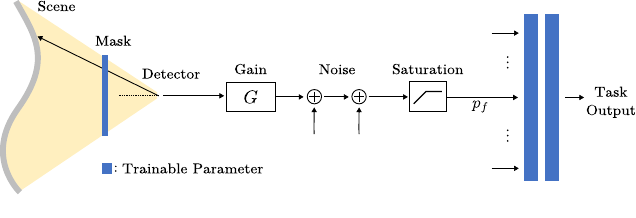}}%
    \put(0.14130982,0.0073727){\color[rgb]{0,0,0}\makebox(0,0)[t]{\smash{\begin{tabular}[t]{c}(a) Optical Effects\end{tabular}}}}%
    \put(0.50308562,0.0086015){\color[rgb]{0,0,0}\makebox(0,0)[t]{\smash{\begin{tabular}[t]{c}(b) Detector Characteristics\end{tabular}}}}%
    \put(0.86036138,0.00983022){\color[rgb]{0,0,0}\makebox(0,0)[t]{\smash{\begin{tabular}[t]{c}(c) Inference Network\end{tabular}}}}%
    \put(0,0){\includegraphics[width=\unitlength,page=2]{sensor-model.pdf}}%
  \end{picture}%
\endgroup%

    \caption{
    \textbf{A minimalist camera as a part of a network.}
    (a)~The optical effects within a freeform pixel include the attenuation due to the mask, the detector’s directional response, and its active area.
    (b)~The detector output is amplified by a gain, degraded by readout and quantization noise, and clipped by the finite dynamic range of the detector.
    (c)~The output $p_f$ of the freeform pixel is fed into the inference network, which uses the outputs of all the pixels of the camera to produce the task output.
    }
    \label{fig:sensor-model}
    \vspace{-0.2in}
\end{figure}

An ideal detector would measure $p_d$, the total light energy that it receives.
A real detector, however, has a gain, noise characteristics, and a finite dynamic range, as illustrated in \cref{fig:sensor-model}.
First, $p_d$ is amplified by a gain $G$ when the detector converts the incident irradiance to an analog signal.
When this analog signal is read out and converted to a digital number, read noise and quantization noise are added, which can be modeled as Gaussian noise ($n_r \sim \mathcal{N}(0, \sigma_r^2)$) and uniform noise ($n_q \sim \mathcal{U}(0,\, p_{lsb})$),\footnote{$p_{lsb}$ is the brightness value corresponding to the least significant bit of the analog-to-digital converter.} respectively. Therefore, the final output of the pixel is:
\begin{equation}
    p_n = G \, p_d + n_r + n_q \,. \label{eq:pn}
\end{equation}
While there are additional sources of noise, such as photon noise and dark current, we only model read and quantization noise since they are the dominant noise sources in our system.
Finally, the detector saturates at a maximum measurement $p_{max}$. Saturation poses a problem during network training because the gradient of a saturated measurement with respect to the trainable parameters in the mask is 0. To avoid such vanishing gradients, we clip the value $p_n$ using a clipping function with a small positive slope in the region of saturation:
\begin{equation}
    p_f = \begin{cases}
        p_n & p_n \le p_{max} \\
        \alpha \, (p_n-p_{max}) + p_{max} & p_n > p_{max}
    \end{cases}, \label{eq:leaky-saturation}
\end{equation}
where $\alpha$ is a small, positive value.

$p_f$ is the final output of a freeform pixel, which serves as an input to the inference network, as shown in \cref{fig:sensor-model}. The sensor model described above has a major impact on the final ``shape'' of a freeform pixel. For instance, a detector's limited dynamic range forces the freeform pixel to ``open up'' to measure enough light to overcome the measurement noise. In the supplemental material, we show that when the above sensor model is not incorporated into the network during the training process, we obtain freeform pixels that perform poorly.

\vspace{-0.12in}
\section{A Toy Example}
\vspace{-0.08in}
\label{sec:sim}

\begin{figure}[t]
    \centering
    \def\svgwidth{\linewidth}
\begingroup%
  \makeatletter%
  \providecommand\color[2][]{%
    \errmessage{(Inkscape) Color is used for the text in Inkscape, but the package 'color.sty' is not loaded}%
    \renewcommand\color[2][]{}%
  }%
  \providecommand\transparent[1]{%
    \errmessage{(Inkscape) Transparency is used (non-zero) for the text in Inkscape, but the package 'transparent.sty' is not loaded}%
    \renewcommand\transparent[1]{}%
  }%
  \providecommand\rotatebox[2]{#2}%
  \newcommand*\fsize{\dimexpr\f@size pt\relax}%
  \newcommand*\lineheight[1]{\fontsize{\fsize}{#1\fsize}\selectfont}%
  \ifx\svgwidth\undefined%
    \setlength{\unitlength}{345.41039658bp}%
    \ifx\svgscale\undefined%
      \relax%
    \else%
      \setlength{\unitlength}{\unitlength * \real{\svgscale}}%
    \fi%
  \else%
    \setlength{\unitlength}{\svgwidth}%
  \fi%
  \global\let\svgwidth\undefined%
  \global\let\svgscale\undefined%
  \makeatother%
  \begin{picture}(1,0.31586924)%
    \lineheight{1}%
    \setlength\tabcolsep{0pt}%
    \put(0,0){\includegraphics[width=\unitlength,page=1]{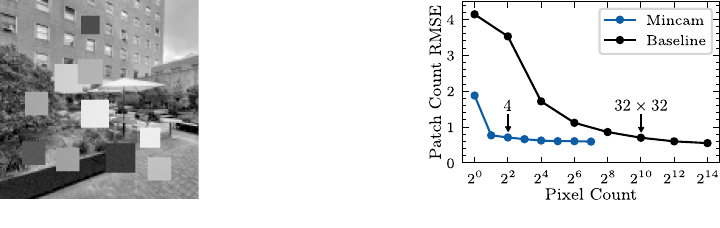}}%
    \put(0.13718462,0.00651397){\color[rgb]{0,0,0}\makebox(0,0)[t]{\lineheight{0}\smash{\begin{tabular}[t]{c}(a) Synthetic Image\end{tabular}}}}%
    \put(0.43546059,0.00651397){\color[rgb]{0,0,0}\makebox(0,0)[t]{\lineheight{0}\smash{\begin{tabular}[t]{c}(b) Freeform Pixels\end{tabular}}}}%
    \put(0,0){\includegraphics[width=\unitlength,page=2]{synthetic-data.pdf}}%
    \put(0.82039592,0.00651397){\color[rgb]{0,0,0}\makebox(0,0)[t]{\lineheight{0}\smash{\begin{tabular}[t]{c}(c) Counting Performance\\\end{tabular}}}}%
    \put(0,0){\includegraphics[width=\unitlength,page=3]{synthetic-data.pdf}}%
  \end{picture}%
\endgroup%

    \vspace{-0.2in}
    \caption{\textbf{Reduction in pixel count with freeform pixels.} 
    (a)~The task is to count the number of patches in an image (up to 10), where the patches have random locations, brightnesses, and sizes. We trained minimalist cameras with an increasing number of freeform pixels (up to 128). 
    (b)~The learned freeform pixels for a 4-pixel minimalist camera.
    (c)~The counting performance of a minimalist camera with these 4 freeform pixels is on par with that of a $32\times32$ baseline camera. This corresponds to a $256\times$ reduction in pixel count. Note that the $x$-axis is scaled logarithmically.}
    \label{fig:synthetic-data}
    \vspace{-0.15in}
\end{figure}

We begin our empirical evaluation of freeform pixels with a synthetic example. The task is to count the number of patches in an image, which is akin to real tasks that involve counting objects such as people or cars. 
\Cref{fig:synthetic-data}(a) shows one such image with $10$ patches.
In each generated image, the number of patches can vary from 0 to 10, and each patch is assigned a random position, brightness, and size (within a range). To simulate occlusion effects, the patches are allowed to partially overlap one another.  Variations in local illumination are simulated by multiplying each image with a smoothly-varying sinusoid with randomly chosen parameters. We synthesized a set of 1,000,000 images for training, 100,000 images for validation, and 250,000 images for testing. 

We trained minimalist cameras (mincams) to count the number of patches, starting with 1 freeform pixel up to 128 freeform pixels, incrementing in powers of 2. The parameters of the sensor model were chosen to be similar to that of a real photodetector. The inference network contains 2 hidden layers, each 128 units wide, with a leaky ReLU as the activation function. The masks of each minimalist camera were initialized with uniform noise, $\mathcal{U}(0.08,\, 0.12)$, and the network was trained by minimizing the cross-entropy loss using the Adam optimizer~\cite{kingmaAdamMethodStochastic2015}.

We compare the performance of the mincams with a traditional camera, where the output of the camera is used as the input to an inference network that is identical in structure to that of the minimalist camera. We refer to this combination of a traditional camera and inference network as the baseline camera. As we lower the resolution of the baseline camera, each pixel simply integrates the light within a larger square. Put another way, the baseline camera can be viewed as a minimalist camera with \textit{fixed} masks, where each mask is a box function.

\Cref{fig:synthetic-data}(b) shows the learned freeform pixels for a minimalist camera with 4 pixels. These freeform pixels achieve 0.71 root-mean-square error (RMSE) in counting patches, on par with the performance of a $32 \times 32$ baseline camera (see \cref{fig:synthetic-data}(c)). This translates to a $256 \times$ reduction in pixel count. This toy example demonstrates that with enough training data, freeform pixels can achieve high performance on a lightweight task using orders of magnitude fewer pixels. 

\vspace{-0.12in}
\section{Camera Architecture}
\vspace{-0.08in}

\begin{figure}[t]
    \centering
    \includegraphics[width=\linewidth]{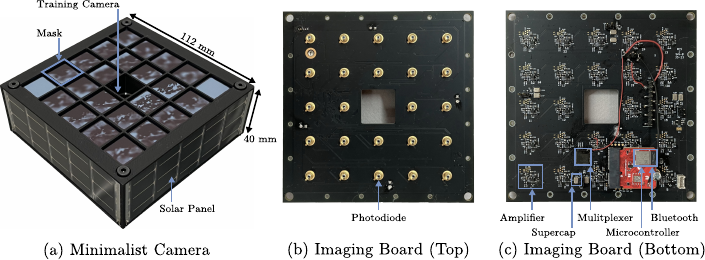}
    \def\svgwidth{\linewidth}
    \vspace{-0.2in}
    \caption{
    \textbf{Hardware prototype of a minimalist camera.} (a) The masks of the pixels are printed on a single transparency, and the corresponding detectors are arranged on the imaging board in (b). (c) The back of the imaging board shows the key components of the camera, including an amplifier for each pixel, a supercap, a multiplexer, and a microcontroller that is Bluetooth enabled. Attached to each side of the camera is a thin solar panel. The energy harvested from the four panels is sufficient for the camera to function in a fully self-powered mode in an indoor environment (see \cref{fig:self-powered}).
    }
    \vspace{-0.2in}
    \label{fig:hardware}
\end{figure}

\Cref{fig:hardware}(a) shows the prototype of the minimalist camera that we have designed and fabricated. It has a total of 24 freeform pixels. The masks of all 24 freeform pixels are printed on a single sheet of transparency film using an inkjet printer. 
The masks can be interchanged by simply sliding a new transparency into a slot in the camera's chassis.\footnote{If a spatial light modulator (SLM) is used in place of the transparency, the masks can be changed via software without any alteration to the hardware.}
Each mask is $16 \times 16 \, \unit{\mm\squared}$ and is placed $11.4 \, \unit{\mm}$ above its detector; this corresponds to a $70\unit{\degree} \times 70\unit{\degree}$ field-of-view for each freeform pixel.

Each detector is a photodiode (Hamamatsu S9119-01), and the array of 24 detectors are arranged on a custom-designed imaging board, the front and back of which are shown in \cref{fig:hardware}(b) and \cref{fig:hardware}(c), respectively. The output of each detector is connected to a transimpedance amplifier which converts the detector's photocurrent to a voltage. The voltages of the 24 freeform pixels are passed through a multiplexer to a microcontroller (STM32WB5MMG), which performs A/D conversion and then wirelessly transmits the measurements to a remote receiver using Bluetooth Low Energy (BLE). A traditional camera (Basler daA1920-160uc) with a  $3\,\unit{\mm}$ lens is attached to the center of the minimalist camera. This camera is only used to capture videos for training the masks of the minimalist camera and to compare its performance with baseline cameras of different resolutions.

Since the minimalist camera generates just a handful of measurements, it consumes very little power during readout and wireless transmission. This allows us to make our prototype completely self-powered. As seen in \cref{fig:hardware}(a), a solar panel (PowerFilm MP3-37) is attached to each of the four sides of the camera to harvest energy from the light falling on it. Since the light incident upon the camera, and hence the harvested energy, can vary over time, the solar panels are connected to an $88 \, \unit{\milli \farad}$ supercapacitor (see \cref{fig:hardware}(c)). In an indoor environment, these panels harvest enough energy to power the camera without using a battery or external power supply.

\begin{figure}[t]
    \centering
    \includegraphics[width=\linewidth]{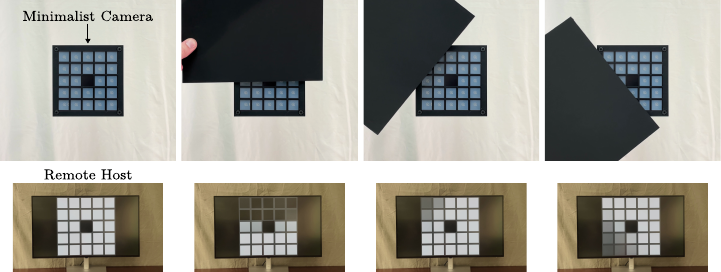}
    \vspace{-0.15in}
    \caption{
    \textbf{Minimalist camera in fully self-powered mode.} The prototype can be entirely powered by just the light falling on it. In a well-lit indoor environment, it can can read out and wirelessly transmit measurements from 24 pixels at 30 frames per second. In this demonstration, the mask of each pixel is uniform in transmittance. A black sheet is moved over the array of pixels, and the wirelessly received measurements are displayed on a remote host shown below. Please see the supplemental video. 
    }
    \label{fig:self-powered}
    \vspace{-0.2in}
\end{figure}

\Cref{fig:self-powered} shows the camera operating in fully self-powered mode. In this demonstration, the ambient illumination falling on each of the camera sides is roughly 
$600$ lux. The camera is able to read out and wirelessly transmit the 24 pixel measurements at 30 frames per second. It can continue to function at lower light levels by simply lowering its framerate. It should be mentioned that the current firmware of the camera is far from optimized. It can be made significantly more power-efficient, enabling the camera to function at much lower light levels. In our lightweight vision experiments, we tethered the camera to a benchtop data acquisition system rather than using the self-powered mode, as this configuration made hardware debugging and synchronization with the training camera easier.

\vspace{-0.1in}
\section{Lightweight Vision: Experiments}
\vspace{-0.05in}
We have used our camera prototype to evaluate the power of freeform pixels in a variety of lightweight vision tasks.

\vspace{-0.1in}
\subsection{Workspace Monitoring}
\label{sec:workspace-monitoring}
\vspace{-0.05in}

\begin{figure}[t]
    \centering
    \includegraphics[width=\linewidth]{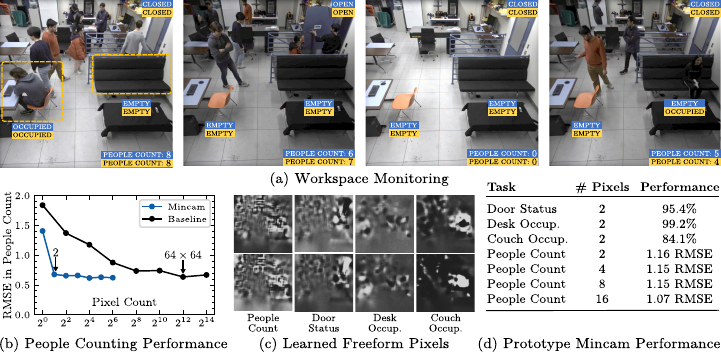}
    \vspace{-0.15in}
    \caption{
    \textbf{Workspace monitoring.}
    (a)~We use a handful of freeform pixels to count the people in the room, determine which zones are occupied (highlighted in yellow boxes in the left image), and detect when the door is open. The outputs of the prototype camera with 8 freeform pixels are shown in blue, and ground truth is shown in yellow.
    (b)~Minimalist cameras and baseline cameras are trained using a labeled video of the scene, and the people counting performance is plotted as a function of pixel count. 
    For this task, the performance of a 2-pixel minimalist camera is close to that of a $64 \times 64$ baseline camera, which corresponds to a $2048\times$ reduction in pixel count.
    (c)~The learned freeform pixels for each task after training the minimalist camera network.
    (d)~The performance of the prototype camera for each of the tasks. 
	}
    \label{fig:workspace-monitoring}
    \vspace{-0.15in}
\end{figure}

In our first application, we use minimalist vision to monitor an indoor space. Consider the workspace shown in \cref{fig:workspace-monitoring}(a). In this scenario, people enter/exit the space, move around, and occupy different zones. Our goal is to monitor the room by the counting of number people in it (from 0 to 8), determining which zones are occupied, and detecting when the door is open. As people move around the space, they occlude each other and different parts of the scene, making each of the above tasks more challenging. Furthermore, over time, the lighting of the space can change dramatically. 
We captured a one-hour video\footnote{No information regarding the identities of individuals in the videos was acquired, stored, or used in the experiments. All of the videos were captured after obtaining signed permissions from the participants.} using the training camera to generate a minimalist camera that can solve all of the above tasks. The video is divided into contiguous segments of 40 minutes for training, 10 minutes for validation, and 10 minutes for testing. 
In each frame of the video, ground truth labels for the tasks are specified.

We generated freeform pixels by training the minimalist camera network, as described in \cref{sec:freeform-pixels}. In \cref{fig:workspace-monitoring}(b), we plot the people counting performance of simulated minimalist cameras and baseline cameras with varying pixel counts. A minimalist camera with 2 freeform pixels achieves {0.68 RMSE} in the number of people, which is comparable to that of a $64\times64$ baseline camera. This translates to a $2048\times$ reduction in pixel count. The masks we used (two for each task) to construct the minimalist camera are shown in \cref{fig:workspace-monitoring}(c). The performance of this camera is seen in the first four rows of the table in \cref{fig:workspace-monitoring}(d). In the four images in \cref{fig:workspace-monitoring}(a), the blue boxes show the  outputs of the system, and the yellow boxes show the ground truth. Also shown in \cref{fig:workspace-monitoring}(d) are people counting performances when we used 4, 8, and 16 freeform pixels.  Please see the supplemental material for a video demonstration of workspace monitoring and the post-processing details.

We now illustrate why a typical minimalist camera does not capture enough visual information to recognize faces. 
State-of-the-art vision systems have attained very high face identification rates (greater than $98\%$) on traditional images~\cite{caoVGGFace2DatasetRecognising2018, dengArcFaceAdditiveAngular2019, raoAttentionAwareDeepReinforcement2017, wangDeepFaceRecognition2021}.
Using 16 freeform pixels specifically designed for counting people, we retrained the inference network to recognize faces on a subset of the CelebA dataset~\cite{liuDeepLearningFace2015}, containing 2751 images of 100 individuals. In this simulation, the faces are scaled to cover the entire field-of-view of the minimalist camera, and each image is augmented with a small amount of noise and a random gain. Once trained, the minimalist camera achieved a recognition rate of $2.0\%$, suggesting that it is unable to perform meaningful face recognition in any real scenario. 
While this does not prove that a minimalist camera guarantees privacy, it strongly supports our conjecture that a person's identity cannot be reliably recovered from the few measurements produced by a minimalist camera.

\vspace{-0.15in}
\subsection{Room Lighting Estimation}
\vspace{-0.05in}
\begin{figure}[t]
    \centering
    \includegraphics[width=\linewidth]{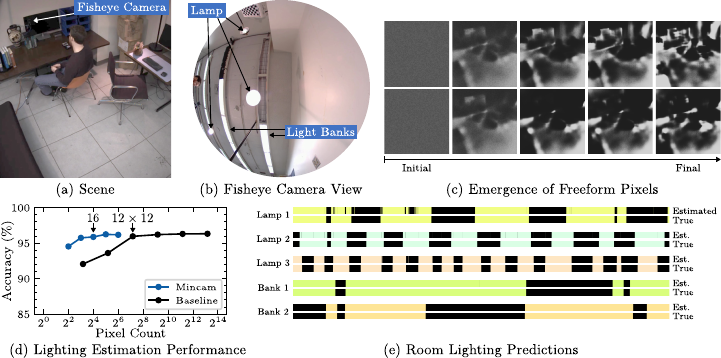}
    \vspace{-0.15in}
    \caption{
    \textbf{Room lighting estimation.}
    (a)~A room lit by three lamps and two overhead light banks. The task is to determine which lights are turned on. A fisheye camera is placed in the scene. 
    (b)~All the lights are visible in the fisheye image, which is used to obtain ground truth labels. 
    We trained a minimalist camera to estimate the room lighting. (c)~The evolution during training of two of the freeform pixels that were initialized with random noise.
    (d) Performance of minimalist cameras and baseline cameras; $16$ freeform pixels are sufficient to achieve the performance of a $12\times12$ baseline camera. (e) The performance of the camera with 8 freeform pixels, compared with ground truth. The black strips correspond to durations for which a light is off.
    }
    \label{fig:lighting}
    \vspace{-0.15in}
\end{figure}

Modern buildings are moving toward optimized lighting systems to reduce their energy consumption, and hence their carbon footprint. In this context, self-sustaining minimalist cameras can be very effective in estimating the ``state'' of the light in a room. Coupled with people counting, a lighting-estimation camera can provide exactly the measurements needed to intelligently optimize lighting. 
Consider the scene shown in \cref{fig:lighting}(a) with three floor lamps and two banks of overhead lights. Our goal is to use minimalist vision to determine the state (on or off) of each of the five lights as people move in and around the space. The lights are not directly visible to the camera. Therefore, the state of the lighting must be inferred from the shading in the scene, even as people move around and obstruct parts of the space.
We captured a 30-minute video of the scene for training and testing. Ground truth labels were obtained using a fisheye camera placed in the scene that directly sees the lights (see \cref{fig:lighting}(b)). 

Using the labeled video, we trained a minimalist camera to determine the state of the room lighting by minimizing the cross-entropy loss for each light. \Cref{fig:lighting}(c) shows the evolution of two freeform pixels during training. Each pixel is initialized to uniform noise, and its shape emerges during the training process.
\Cref{fig:lighting}(d) compares the performance of minimalist cameras with baseline cameras; a $12\times12$ baseline camera is needed to achieve the same performance as a minimalist camera with $16$ freeform pixels.
We fabricated 8 freeform pixels, which could estimate the room lighting (the states of all five lights) with $94.0\%$ accuracy. A comparison between the outputs of the camera and the ground truth is shown in \cref{fig:lighting}(e). Please see the supplemental video.

\vspace{-0.15in}
\subsection{Traffic Monitoring}
\vspace{-0.05in}

\begin{figure}[t]
    \centering
    \includegraphics[width=\linewidth]{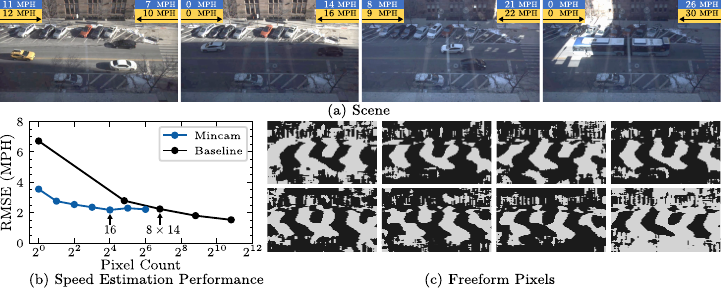}
    \vspace{-0.2in}
    \caption{
    \textbf{Traffic speed estimation. }
    (a)~The task is to estimate the average traffic speed in both the left and right directions. A video of the scene captured over an entire day was used to train minimalist cameras.
    (b)~The performance of minimalist and baseline cameras. A minimalist camera with 16 freeform pixels achieves the same performance as an $8 \times 14$ baseline camera.
    (c)~We fabricated a minimalist camera with 8 freeform pixels, which can monitor traffic speed with $2.30$ RMSE in miles per hour. In (a), the outputs of the camera are shown in blue, and the ground truth in yellow.
    }
    \vspace{-0.15in}
    \label{fig:traffic}
\end{figure}

Minimalist cameras can be attached, without cables or external power, to poles or buildings to monitor traffic. In \cref{fig:traffic}(a), the task is to estimate the average traffic speed in both directions (left and right). In this system, the minimalist camera uses the temporal history of its pixel measurements over a period of one second to perform the task. The inference network outputs two values: the left and right traffic speeds, in miles per hour.
We collected training data by capturing a video of the scene over an entire day and randomly extracted five-minute video clips for validation and testing. The ground truth labels were obtained by applying an off-the-shelf object detector~\cite{jocherUltralyticsYOLOv82023} to the captured video to track individual cars. The network was trained by minimizing the mean squared error between the predicted and ground truth traffic speeds. 

\Cref{fig:traffic}(b) compares the performance of minimalist cameras with that of baseline cameras.
We fabricated the 8 freeform pixels shown in \cref{fig:traffic}(c), which were able to estimate the left and right traffic speeds with an RMSE of $2.30$ miles per hour. Please see the supplemental material for a video of traffic monitoring and details of the network training and post-processing.

\vspace{-0.05in}
\section{Discussion}
\vspace{-0.05in}

We have introduced the concept of freeform pixels and shown how they can be effective in solving lightweight vision tasks using just a handful of measurements. There are several directions in which we plan to extend our work. First, in place of the printed transparency we used for our optical masks, a spatial light modulator, such as a liquid-crystal display, can be used to set the shapes of the masks electronically. This would allow us to change the functionality of the minimalist camera as a function of time. This also implies that different tasks can be time-multiplexed, allowing us to use more freeform pixels for any given task. In addition, spatio-temporal control of the masks would allow us to extract more revealing visual features, particularly in the case of dynamic scenes.

While our current notion of a freeform pixel performs a linear projection of the scene, we are interested in generalizing the concept so that it can perform more advanced optical mappings. For instance, by using lenses in addition to a mask, each pixel can be designed to apply a convolution to the scene with a pre-trained kernel. Such a system can also be modeled as a part of a network and has the potential to solve more sophisticated tasks.

With the above enhancements, we believe minimalist cameras can be designed to perform a wider range of vision tasks, while still guaranteeing privacy protection and self-sustainability. Ultimately, our goal is to use minimalist vision to address existing needs in the fields of environment sensing, wildlife monitoring, crowd and traffic analysis, and energy conservation.

\section*{Acknowledgements}
\vspace{-0.05in}
This work was supported by the Office of Naval Research (ONR) under award number N00014-21-1-2378. The authors are grateful to Behzad Kamgar-Parsi for his support and encouragement. The authors also thank Carl Vondrick for his technical feedback all through the project and Mikhail Fridberg for his help with designing the camera electronics.

%
%
\bibliographystyle{splncs04}
\bibliography{Mincam}

\setlength{\tabcolsep}{6pt}
\renewcommand{\arraystretch}{1.1}

\renewcommand\thesection{\Alph{section}}
\renewcommand\thesubsection{\thesection.\arabic{subsection}}

\newpage
{
\centering
\Large
\textbf{Minimalist Vision with Freeform Pixels:\\Supplementary Material} \\
\vspace{1.0em}
}
\appendix

\section{Importance of the Sensor Model}
Here we examine the effect of training freeform pixels without including the sensor model in the network. 
Once trained, we evaluate the freeform pixels in a simulated minimalist camera that includes the optical effects and detector characteristics of the sensor model.

Using the synthetic dataset for counting patches described in Sec. 4 of the main paper, we generated 4 freeform pixels by training a minimalist camera without the sensor model. 
We then froze the learned masks and retrained the inference network with the sensor model included in the network. The parameters of the sensor model were chosen to be similar to that of our hardware prototype. After retraining the inference network, the 4 freeform pixels achieved 2.28 root-mean-square (RMS) error in the number of patches.
By comparison, 4 freeform pixels that were trained for counting patches with the same sensor model incorporated in the network during training achieved 0.93 RMS error in the number of patches.
This performance gap between the two minimalist cameras demonstrates that including the sensor model in the network during the training process is critical to generate performant freeform pixels.

\section{Camera Architecture Details}
\Cref{tab:parts} lists the components used in our prototype minimalist camera. Each detector is connected to a transimpedance amplifier with a gain of $10^7\,\unit{\V / \A}$. 
In the lightweight vision experiments, we used a National Instruments USB-6363 data acquisition unit to simultaneously read out the freeform pixel measurements and trigger the training camera.
Since the detectors and training camera are sensitive to near-infrared wavelengths, we mounted a filter in front of the minimalist camera to block near-infrared light.

The sensor model parameters corresponding to the hardware prototype were either empirically measured or extracted from component datasheets.
First, the detector datasheet~\cite{hamamatsuDiode} specifies the active area to be $0.88 \times 0.88 \, \unit{\mm}^2$ and publishes the directional response.
We use $\sigma_r = 400\, \unit{\mu\V}$ as the standard deviation of the read noise. 
Quantization noise and sensor saturation are based on a $16$-bit detector that saturates at $3.2\,\unit{\V}$.
Finally, our process of printing masks on a transparency can only fabricate masks with transmittance values in the range $0.01 \le M(x,y) \le 0.67$. We account for this fabrication limitation by scaling the mask transmittance values to this range during the training process. 

\begin{table}[t]
    \caption{\textbf{List of components in the prototype minimalist camera.}}
    \centering
    \begin{tabular}{lcc}
        \toprule
        Component & Quantity & Description \\
        \midrule
        Detector & 24 & Hamamatsu S9119-01 \\
        Amplifier & 24 & TLV521DCKR \\
        Multiplexer & 1 & ADG732BCPZ \\
        Microcontroller & 1 & STM32WB5MMG \\
        Photovoltaic & 4 & PowerFilm MP3-37 \\
        Supercapacitor & 8 & $11\,\unit{\milli\farad}$, each \\
        Training Camera & 1 & Basler daA1920-160uc \\
        Training Camera Lens & 1 & Edmund Optics $3\,\unit{\mm}$, $f/2.5$ \\
        Infrared Filter & 1 & Schott KG3 \\
        \bottomrule
    \end{tabular}
    \label{tab:parts}
    \vspace{-0.1in}
\end{table}

\section{Lightweight Vision Experimental Details}

Slight mismatches between the sensor model and hardware prototype cause deviations between the simulated and real measurements of each freeform pixel. Furthermore, radiometric and geometric misalignments between the freeform pixels and training camera contribute to this mismatch.
To account for this mismatch after the freeform pixels are fabricated, we retrain the inference network using pairs of real measurements generated by the prototype and their corresponding ground truth labels.
This processes necessitates the capture of two datasets for each lightweight vision experiment.
The first dataset contains a training video that is only used to the generate masks of the freeform pixels that will be fabricated.
Once the masks are fabricated, a dataset is captured containing a video from the training camera and corresponding measurements from the freeform pixels. This dataset, which is summarized in \cref{tab:dataset-sizes} for each task, is used to retrain the inference network of the hardware prototype and train simulated minimalist cameras and baseline cameras.

\subsection{Workspace Monitoring}
The networks for counting people were trained by minimizing the mean squared error between the predicted and ground truth people count.
The networks for the remaining tasks (detecting the state of the door and occupancy of the zones) were trained by minimizing the cross-entropy loss.
At test time, the predicted people count from both the baseline and minimalist cameras is rounded to the nearest integer and then filtered using a 2-second median filter. In the supplemental video, the outputs produced by the minimalist camera for detecting the state of the door and the zone occupancy are filtered using a $0.5$-second median filter.

We fabricated 16 freeform pixels for counting people. We then used a greedy algorithm to iteratively remove the least important pixels from the collection to evaluate the counting performance using a smaller number of pixels. At each iteration in this algorithm, the ``least important'' freeform pixel is the one which, when removed from the collection, admits the smallest increase in validation loss. Figure 7(d) in the main paper shows the performance of subsets of the 16 freeform pixels obtained using this approach.

As explained in the main paper, we generated a dataset to evaluate the face identification performance of the 16 freeform pixels designed for counting people. We constructed the dataset using 100 randomly chosen identities in the CelebA dataset~\cite{liuDeepLearningFace2015} that each appear in at least 20 images. The training, validation, and testing sets are composed such that each set contain images of all 100 individuals. We trained minimalist camera networks to convergence by performing a grid search over the batch size, learning rate, and the inference network's width and depth.

\begin{table}[t]
    \caption{\textbf{Sizes of the datasets used in the lightweight vision experiments.}}
    \centering
    \begin{tabular}{lccc}
        \toprule
        Experiment & Dataset Split & Duration (min.) & \# Samples \\
        \midrule
        \multirow{3}{*}{Workspace Monitoring}
        & Training & 38 & 68,069 \\
        & Validation & 11 & 19,400 \\
        & Testing & 10 & 18,720 \\
        \midrule
        \multirow{3}{*}{Lighting Estimation}
        & Training & 17 & 31,411 \\
        & Validation & 6 & 10,162 \\
        & Testing & 6 & 10,738 \\
        \midrule
        \multirow{3}{*}{Traffic Monitoring}
        & Training & 166 & 23,951 \\
        & Validation & 21 & 3,479 \\
        & Testing & 42 & 7,118 \\
        \bottomrule
    \end{tabular}
    \label{tab:dataset-sizes}
    \vspace{-0.1in}
\end{table}

\subsection{Traffic Monitoring}
Both the minimalist camera and baseline camera use the temporal history of measurements over a period of one second (a stack of 30 measurements) to estimate the average traffic speeds.
We apply forward differencing in the time domain to the measurement stack before passing it through the inference network.
We found empirically that applying forward differencing improved the performance of both the baseline and minimalist camera networks.

The validation and test sets for traffic monitoring are extracted by randomly sampling five-minute clips from the eight-hour video, as described in the main paper. The remaining portions of the video are used for training.
The datasets are generated by extracting overlapping one-second periods from the video.
Some clips that do not contain any traffic are removed from the datasets.
To retrain the inference network of the hardware prototype, we generated a larger number of training and validation samples (142,933 and 20,770, respectively) by extracting one-second periods with more aggressive overlap.

At test time, the estimated traffic speeds from both the minimalist camera and baseline camera are filtered using a 2-second median filter. We also observed that the object detector used for ground truth labeling is only accurate within a field-of-view that is slightly smaller than that of the baseline and minimalist camera. This caused labeling errors when a moving vehicle appears near the edge of the image. To minimize the effect of this labeling error on the computed performance (i.e. the RMS error of the predicted traffic speeds), we set the predicted speed of both the minimalist and baseline cameras to 0 when the ground truth speed is less than 3 miles per hour.

%
%
\bibliographystyle{splncs04}
\bibliography{Mincam}

\end{document}